\documentclass[conference]{IEEEtran}
\IEEEoverridecommandlockouts
\usepackage{cite}
\usepackage{amsmath,amssymb,amsfonts}
\usepackage{algorithmic}
\usepackage{graphicx}
\usepackage{textcomp}
\usepackage{xcolor}

\usepackage{booktabs}
\usepackage{graphicx}
\usepackage{microtype}
\makeatletter
\newcommand{\linebreakand}{%
  \end{@IEEEauthorhalign}
  \hfill\mbox{}\par
  \mbox{}\hfill\begin{@IEEEauthorhalign}}
\makeatother

\def\BibTeX{{\rm B\kern-.05em{\sc i\kern-.025em b}\kern-.08em
    T\kern-.1667em\lower.7ex\hbox{E}\kern-.125emX}}
\begin{document}

\title{GHR-VLM: Making Zero-Shot Transit Video Analytics Realizable with Grounded Hybrid Reasoning}

\author{\IEEEauthorblockN{Kaicong Huang}
\IEEEauthorblockA{
%\textit{Civil and Environmental Engineering} \\
\textit{Rensselaer Polytechnic Institute}\\
Troy, NY, USA \\
huangk10@rpi.edu}
\and
\IEEEauthorblockN{Weiheng Oh}
\IEEEauthorblockA{
%\textit{Civil and Environmental Engineering} \\
\textit{Rensselaer Polytechnic Institute}\\
Troy, NY, USA \\
ohw@rpi.edu}
\and
\IEEEauthorblockN{Jack M. Reilly}
\IEEEauthorblockA{
%\textit{Civil and Environmental Engineering} \\
\textit{Rensselaer Polytechnic Institute}\\
Troy, NY, USA \\
REILLJ2@rpi.edu}
\linebreakand
\IEEEauthorblockN{Thomas Guggisberg}
\IEEEauthorblockA{
%\textit{Civil and Environmental Engineering} \\
\textit{Capital District Transportation Authority}\\
Albany, NY, USA \\
thomas@cdta.org}
\and
\IEEEauthorblockN{Ruimin Ke$^*$\thanks{$^*$ Corresponding author.}}
\IEEEauthorblockA{
%\textit{Civil and Environmental Engineering} \\
\textit{Rensselaer Polytechnic Institute}\\
Troy, NY, USA \\
ker@rpi.edu}
}

% Full-width overview figure placed directly below the author block.
\makeatletter
\IEEEaftertitletext{%
  \begin{center}
    \includegraphics[width=\textwidth]{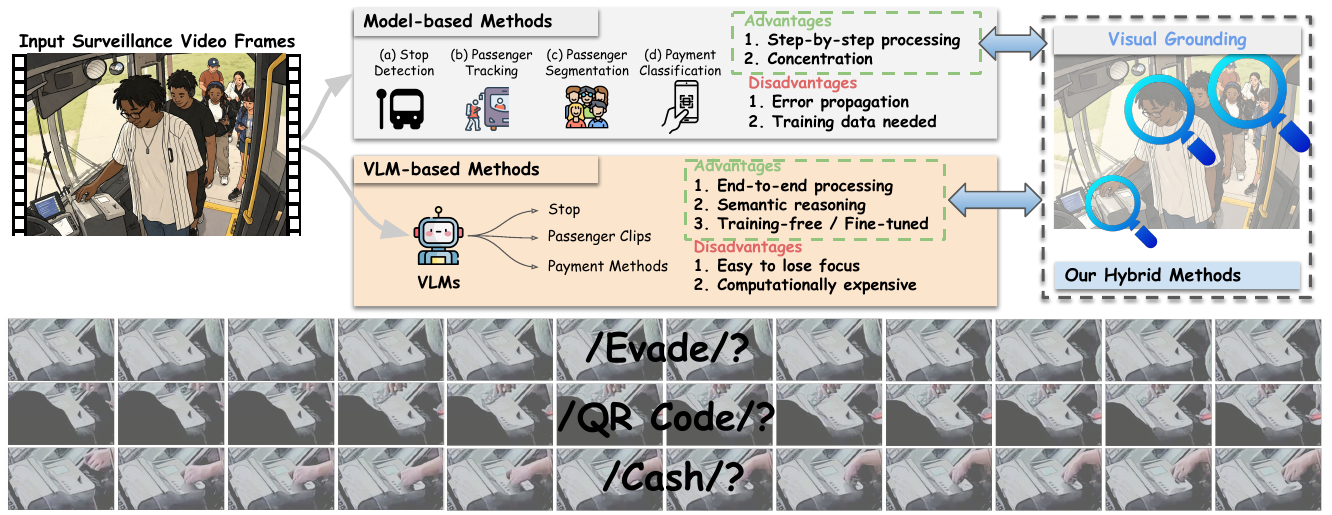}
  \end{center}
  \refstepcounter{figure}%
  \@makecaption{\figurename~\thefigure}{Comparison of model-based, VLM-based, and our grounded hybrid approaches for transit video analytics. Model-based pipelines provide explicit visual grounding but require task-specific training and suffer from error propagation. VLM-based methods offer open-vocabulary reasoning but may lose focus in long videos and incur high inference costs. Our hybrid approach integrates grounded stepwise processing with selective VLM reasoning, enabling zero-shot transit video analytics.}%
  \label{fig:intro}
  \vspace{1ex}%
}
\makeatother

\maketitle

\begin{abstract}
Transit video understanding can provide valuable fine-grained data that conventional passenger counters and fare systems cannot capture. However, supervised video models require task-specific annotations, while applying vision-language models (VLMs) directly to long onboard videos is unreliable and costly.
To leverage the complementary strengths of both approaches, we propose GHR-VLM, a visual grounded hybrid reasoning framework for zero-shot transit-bus video analytics. It is motivated by the observation that explicit visual grounding can improve VLM reasoning by converting long surveillance streams into compact, passenger-centered spatiotemporal evidence. 
Specifically, we propose an edge-cloud design in which a lightweight edge-based monitor continuously tracks door status and segments passenger clips. A backend VLM then identifies boarding passengers and classifies payment behavior through a two-stage coarse-to-fine refinement of spatiotemporal evidence.
By invoking the VLM only on grounded passenger clips and contact sheets, GHR-VLM reduces cloud inference, avoids payment-specific training data, and supplies the localized evidence that VLMs otherwise struggle to identify.
Evaluation on 486 minutes of real-world bus surveillance video demonstrates the potential of grounded edge-cloud reasoning for passenger-level payment analytics while highlighting the challenges posed by degraded video conditions.
\end{abstract}

\begin{IEEEkeywords}
transit video analytics, vision-language models, visual grounding, edge-cloud collaboration, zero-shot recognition
\end{IEEEkeywords}

\section{Introduction}

Transit agencies require fine-grained operational data for service planning, crowding management, revenue auditing, and passenger-experience assessment~\cite{huang2025transitreid}. Existing passenger counters, fare logs, and inspection records provide only partial evidence and cannot associate each stop event with individual passenger activities~\cite{dib2023unified,jahn2022engineering,correa2017fare}. Automated payment analytics must therefore identify boarding and alighting passengers, localize payment events, recognize payment methods, and detect unpaid boarding.

This problem involves vehicle-state detection, passenger tracking, temporal segmentation, activity understanding, and fine-grained hand-object recognition. Existing fare-evasion methods learn specific behaviors from annotated keypoints or action examples~\cite{van2024deep}, while more general video models acquire spatiotemporal representations through supervised training~\cite{yan2018spatial,huang2026ipay,feichtenhofer2019slowfast,tong2022videomae,bertasius2021space,arnab2021vivit}. Their reliance on task-specific labeled data, however, limits their applicability to onboard payment analytics, where payment behaviors are \textbf{diverse, sparsely observed, and costly to annotate}. We therefore formulate the problem as a \textbf{zero-shot fine-grained spatiotemporal task}, in which brief farebox interactions must be localized and associated with the correct passenger in low-resolution, cluttered video.

Video vision-language models (VLMs) offer a promising language-driven interface for such difficult-to-annotate events. Image- and video-language models have demonstrated open-vocabulary recognition, visual grounding, instruction following, multimodal reasoning, and temporal understanding~\cite{li2023blip,liu2023visual,bai2023versatile,wang2021actionclip,maaz2024video,zhang2023video,song2024moviechat,lin2024video,bai2025qwen3}. However, directly prompting VLMs on long onboard videos is unreliable and expensive. Spatially, blur, poor illumination, low resolution, and occlusion obscure the localized cues that distinguish payment methods. Temporally, payment actions occupy only a small fraction of the stream and require precise segmentation. Fleet-scale cloud inference also conflicts with edge constraints on latency, bandwidth, and computation~\cite{ren2023survey}.

These limitations motivate us to combine VLM reasoning with explicit spatiotemporal grounding. Detectors, segmentation models, and trackers can localize passengers, associate trajectories, and extract relevant intervals~\cite{carion2020end,wojke2017simple,zhang2022bytetrack}, while open-vocabulary grounding and promptable segmentation reduce the annotation effort required for deployment~\cite{li2022grounded,minderer2022simple,liu2024grounding,kirillov2023segment,ravi2025sam,ren2024grounded}. Together, these components can transform long surveillance streams into compact, passenger-centered evidence for VLM reasoning.

We propose GHR-VLM, a Visual Grounded Hybrid Reasoning framework for zero-shot transit-bus video analytics. Lightweight edge modules identify the front-door and payment regions, extract stop intervals from door states, and detect, track, and separate passengers within those intervals. A backend VLM then distinguishes boarding passengers from alighting or already-onboard passengers and classifies payment behavior through a two-stage coarse-to-fine refinement of spatiotemporal farebox evidence. By invoking the VLM only on grounded clips and contact sheets, GHR-VLM reduces cloud inference, avoids task-specific payment training data, and supplies the precise evidence that VLMs otherwise struggle to localize.

This paper makes the following contributions:
\begin{itemize}
\item We introduce a grounded hybrid reasoning framework for zero-shot transit video analytics, in which model-based perception provides structured spatiotemporal evidence to guide the open-vocabulary semantic reasoning of VLMs.
\item We develop an end-to-end bus payment data collection pipeline that transforms a single onboard surveillance stream into structured stop-level, passenger-level, and payment-level events.
\item We instantiate an edge-cloud collaborative architecture designed for resource-constrained in-vehicle deployment.
\end{itemize}

% \section{Related Work}

\section{Transit Data Collection}

Transit data collection is typically divided into passenger counting, transaction logging, and inspection planning. Learned counters improve boarding and alighting estimates but remain focused on aggregate counts~\cite{jahn2022engineering}. Fare-card and farebox records provide transaction data, yet they contain blind spots such as unrecorded evasion and incomplete deployment, motivating their fusion with passenger counts~\cite{dib2023unified}. Operations research further addresses fare evasion through network-level inspection and deterrence models~\cite{correa2017fare}. However, none of these approaches reconstructs a passenger-level visual chain from boarding to payment. Vision-based methods address parts of this problem through supervised fare-evasion and action recognition~\cite{van2024deep,huang2026ipay,yan2018spatial,feichtenhofer2019slowfast,tong2022videomae,bertasius2021space,arnab2021vivit}. Although effective with labeled data and stable scene geometry, they require task-specific training and usually predict only narrow fraud categories rather than complete payment events. Moreover, onboard buses are less constrained than stations or gates. Passengers enter in groups, occlude one another, use diverse payment media, and interact with small farebox components under poor lighting. GHR-VLM therefore formulates transit data collection as grounded video understanding rather than isolated counting or fare-evasion classification.

\section{Methodology}

\begin{figure*}[t]
  \centering
  \includegraphics[width=\textwidth]{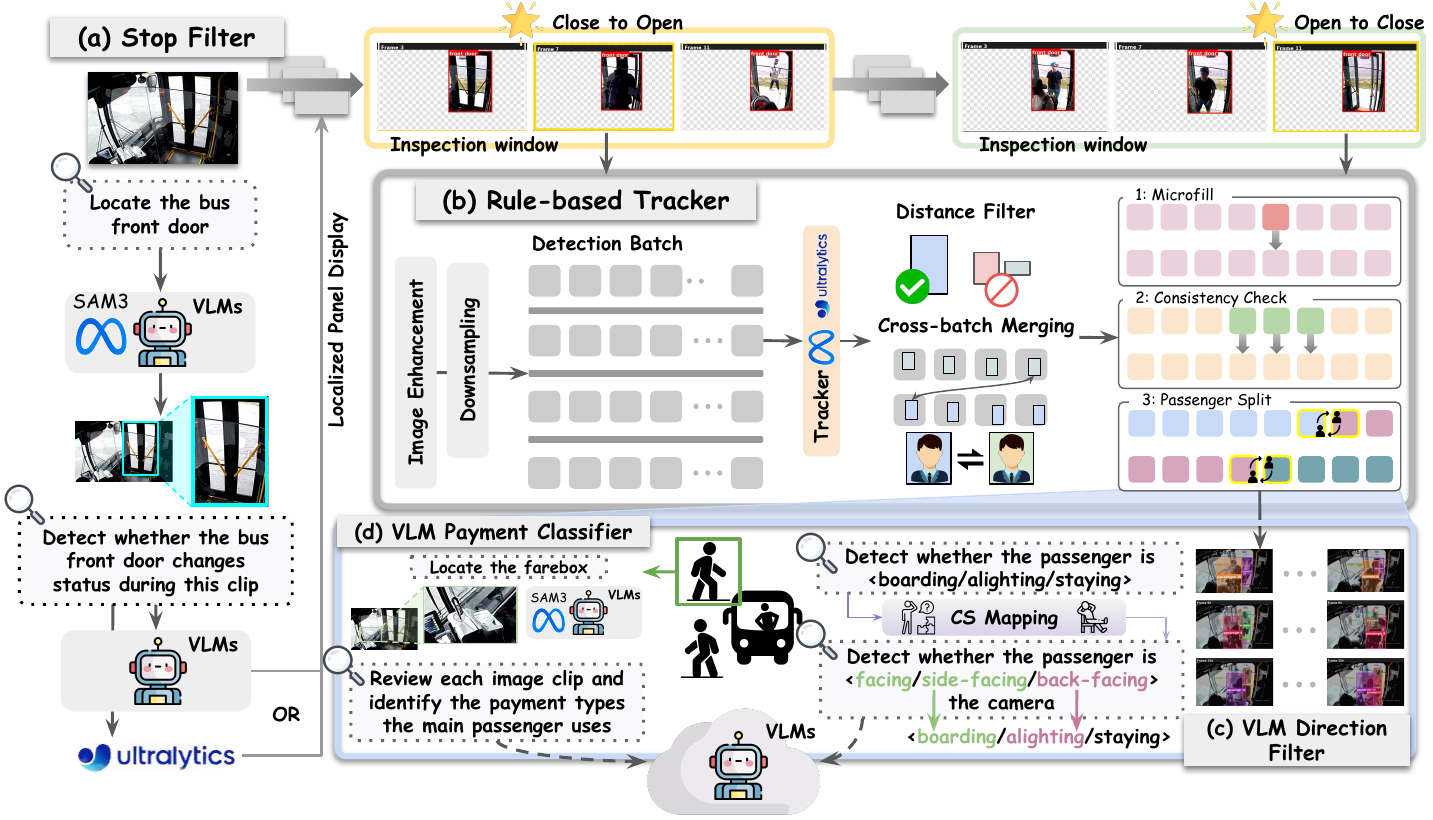}
  \caption{Overview of GHR-VLM. The edge pipeline grounds the front-door region and restricts downstream processing to stop intervals during which the door is open. A rule-based tracker then generates passenger-specific clips, a VLM-based direction filter retains only boarding passengers, and a farebox-grounded two-stage VLM assigns payment labels.}
  \label{fig-framework}
\end{figure*}

Given a continuous onboard video $\mathcal{V}=\{I_t\mid 0\leq t\leq T\}$, GHR-VLM produces passenger-level transit records
\begin{equation}
    \mathcal{R}=\bigl\{(S_k,P_{k,j},\hat{a}_{k,j},\hat{y}_{k,j})\bigr\}_{k,j}
\end{equation}
where $I_t$ is the frame at timestamp $t$, $T$ is the video duration, $S_k$ is the $k$-th stop interval, $P_{k,j}$ is the $j$-th passenger clip within that stop, $\hat{a}_{k,j}$ is the passenger direction, and $\hat{y}_{k,j}$ is the payment label. The label space is $\mathcal{Y}=\{\text{QR},\text{cash},\text{tap},\text{swipe},\text{evade}\}$. Lightweight and promptable vision modules continuously localize relevant evidence, while VLM inference is reserved for compact grounded inputs. Figure~\ref{fig-framework} shows the four stages described below.

\subsection{Door-Grounded Stop Filtering}

Most frames in a continuous trip contain no passenger exchange, so GHR-VLM uses the front door to identify likely stops. SAM 3~\cite{carion2025sam} localizes the front entrance in the first processed frame and returns a normalized box $B_d$, which is cached for the full video under the fixed-camera setting. The box is expanded by 20\% to retain the doorway context, while pixels outside it are masked to suppress irrelevant motion.

The VLM analyzes the cropped stream in overlapping 10-second windows, each represented by 12 uniformly sampled frames. It predicts \emph{no change}, \emph{closed-to-open}, or \emph{open-to-closed}, and identifies the first frame showing the new state. A second VLM call verifies each candidate within a five-second window on either side of its timestamp, and duplicate transitions within three seconds are merged. The verified pins are $\mathcal{P}_d=\{(\hat{t}_\ell,\hat{z}_\ell)\}_{\ell=1}^{L}$, where $L$ is the number of retained pins, $\hat{t}_\ell$ is a timestamp, and $\hat{z}_\ell\in\{c\!\rightarrow\!o,o\!\rightarrow\!c\}$ is the transition type.

Let $t_k^o$ denote the timestamp of the $k$-th opening pin, and let $\mathcal{C}_k$ contain all verified closing timestamps after $t_k^o$. The matched closing time is
\begin{equation}
t_k^c=\min \mathcal{C}_k
\end{equation}

When $\mathcal{C}_k$ is empty, the processing endpoint is used as $t_k^c$. The resulting stop interval is
\begin{equation}
S_k=[t_k^o,t_k^c]
\end{equation}

Only frames within these intervals enter passenger tracking. We also provide a lightweight alternative that applies a fine-tuned YOLO11~\cite{jocher2024ultralytics} to each grounded door crop.

\subsection{Rule-Based Passenger Tracking and Temporal Segmentation}

Each stop interval $S_k$ defines the temporal range for passenger tracking. The system then processes alternate frames and enhances interior visibility through adaptive contrast, shadow amplification, and highlight suppression. SAM 3 tracks instances prompted as ``passenger'' in 16-frame batches, while boxes covering less than 5\% of the image or wider than tall are discarded.

Batching reduces edge memory usage but can reset identities across batches. We therefore match each detection in the first frame of a new batch to the best unmatched detection in the final frame of the previous batch. Their overlap is
\begin{equation}
    \operatorname{IoU}(B,B')=\frac{|B\cap B'|}{|B\cup B'|}
\end{equation}
where $B$ is a boundary box from the preceding batch and $B'$ is a boundary box from the new batch. A match is accepted when the IoU exceeds $\theta_{\mathrm{I}}=0.20$. The matched local identity inherits the corresponding global identity.

To identify the primary passenger in each frame, we exploit the fixed camera geometry as a projective prior, assuming that the passenger closest to the fare area is the target.
Let $\mathcal{D}_{k,i}$ be the valid tracked identities visible in sampled frame $i$ of stop $k$, and let $B_{i,j}$ be the box of identity $j$. The main identity is selected by
\begin{equation}
    m_i=\arg\max_{j\in\mathcal{D}_{k,i}}\operatorname{area}(B_{i,j})
\end{equation}
The index $i\in\{1,\ldots,N_k\}$ identifies one of the $N_k$ sampled frames. The value $m_i$ is set to $-1$ when $\mathcal{D}_{k,i}$ is empty.

To mitigate brief identity switches under poor visual conditions, we apply a two-step repair that converts the raw sequence ${m_i}$ into $\tilde{\mathbf{m}}_k$. First, a one-frame identity differing from both neighbors is replaced by the preceding identity. Second, a valid run of up to three frames is replaced by the preceding identity, while a missing run of the same length is filled only when its neighboring identities agree. These steps correspond to the microfill and consistency operations in Figure~\ref{fig-framework}.

Let $R_{k,j}$ be the $j$-th maximal run with one valid repaired identity $q_{k,j}\geq0$. Its sampled-frame extent is
\begin{equation}
    R_{k,j}=[s_{k,j},e_{k,j}]=\operatorname{MaxRun}_j(\tilde{\mathbf{m}}_k)
\end{equation}
where $s_{k,j}$ and $e_{k,j}$ are the first and last sampled-frame indices in the run. Let $\tau_{k,i}$ be the absolute timestamp of sampled frame $i$. The clip begins at $\tau_{k,s_{k,j}}$ and ends at the timestamp of the next sampled frame. A run reaching the last sampled frame ends at $t_k^c$. Denoting these boundaries by $\tau_{k,j}^{s}$ and $\tau_{k,j}^{e}$ gives
\begin{equation}
    P_{k,j}=\{I_t\mid \tau_{k,j}^{s}\leq t<\tau_{k,j}^{e}\}
\end{equation}
Thus, the repaired identity sequence determines a maximal run, the run determines temporal boundaries, and the boundaries determine the passenger clip. Adjacent run boundaries within three sampled frames are merged, while clips shorter than 0.5 seconds are removed. The retained clips become the inputs to passenger direction filtering.

\subsection{Complex-to-Simple Passenger Direction Mapping}

We assume that only boarding passengers proceed to fare payment, so non-boarding passengers are filtered out. Direct activity recognition, however, requires transit-specific spatiotemporal knowledge that a general-purpose VLM may lack. We therefore introduce a Complex-to-Simple (CS) mapping that reformulates the task as a generic facing-direction judgment.

Each candidate clip $P_{k,j}$ contains one temporally isolated passenger. We uniformly sample $M=4$ chronological frames to form $\mathcal{F}{k,j}$ and submit only this sequence to the VLM. Given the orientation-focused prompt $q_{\mathrm{face}}$, the VLM produces
\begin{equation}
    \hat{o}_{k,j}=\Phi_{\mathrm{face}}(\mathcal{F}_{k,j},q_{\mathrm{face}})
\end{equation}
where $\Phi_{\mathrm{face}}$ denotes the VLM for this task, and $\hat{o}_{k,j}$ is its visual-state judgment from $\mathcal{O}=\{\mathrm{front},\mathrm{side},\mathrm{back},\mathrm{inside}\}$. A fixed rule $\psi_{\mathrm{CS}}$ then maps this generic judgment to passenger direction
\begin{equation}
    \hat{a}_{k,j}=\psi_{\mathrm{CS}}(\hat{o}_{k,j})=
    \begin{cases}
        \mathrm{boarding}, & \hat{o}_{k,j}\in
        \{\mathrm{front},\mathrm{side}\},\\
        \mathrm{alighting}, & \hat{o}_{k,j}=\mathrm{back},\\
        \mathrm{staying}, & \hat{o}_{k,j}=\mathrm{inside}
    \end{cases}
\end{equation}
This rule injects the fixed inward-facing camera geometry as domain knowledge, so the VLM only judges visual orientation. The boarding gate is
\begin{equation}
    g_{k,j}=\mathbb{I}[\hat{a}_{k,j}=\mathrm{boarding}]
\end{equation}
where $\mathbb{I}$ is the indicator function. Only clips with $g_{k,j}=1$ proceed to payment classification.

\begin{figure*}[t]
  \centering
  \includegraphics[width=\textwidth]{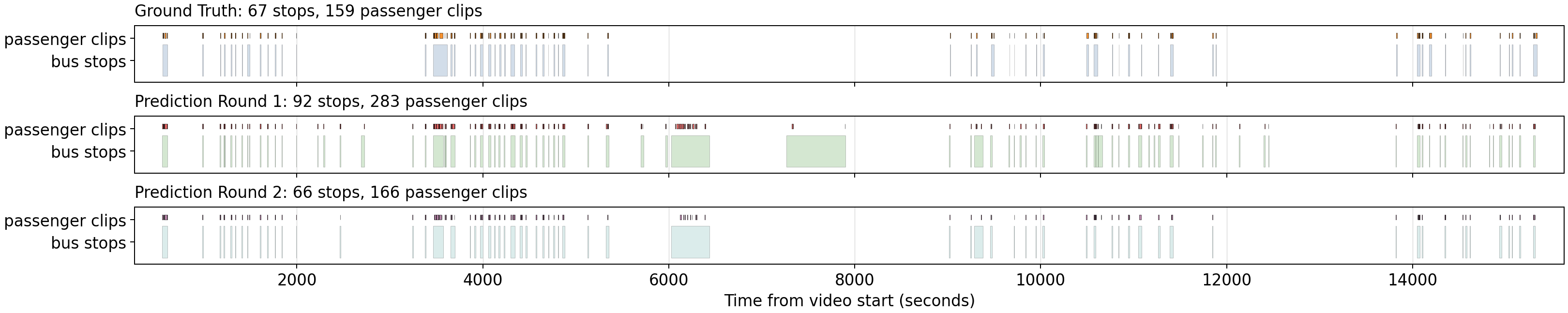}
  \vspace{1mm}
  \includegraphics[width=\textwidth]{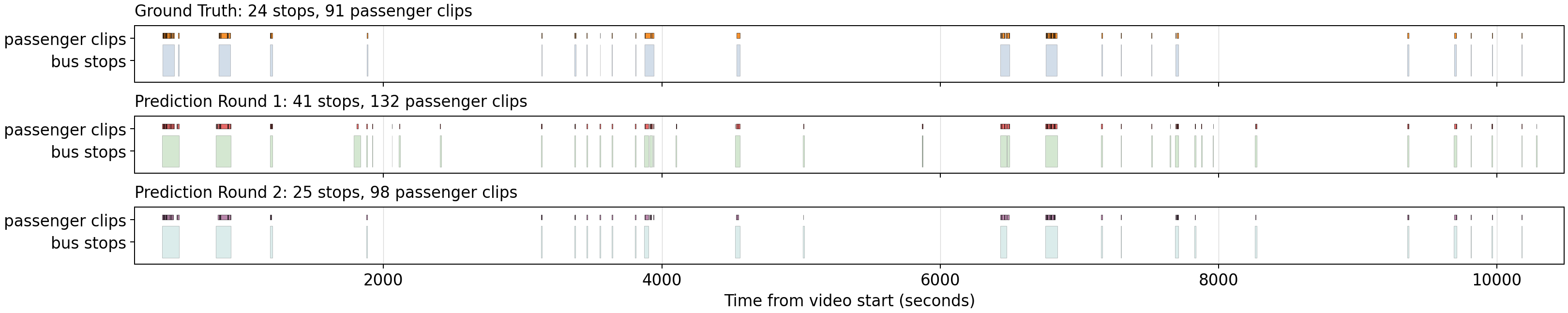}
  \caption{Full-video timelines of detected stops and passenger clips for C3\_1 (top) and C3\_3 (bottom). Each panel compares the ground truth with the raw Round 1 predictions and the refined Round 2 predictions after direction filtering and stop merging.}
  \label{fig-edge-timelines}
\end{figure*}

\begin{table*}[t]
\centering
\caption{Stop- and passenger-clip interval detection using one-to-one matching at temporal IoU $\geq 0.2$. The best precision (Prec.), recall (Rec.), and F1 score for each video are shown in bold.}
\label{tab-edge-results}
\small
\renewcommand{\arraystretch}{0.8}
\setlength{\tabcolsep}{7pt}
\begin{tabular}{@{}lcccccccc@{}}
\toprule
& \multicolumn{2}{c}{\textbf{Predicted Count}}
& \multicolumn{3}{c}{\textbf{Stop Intervals}}
& \multicolumn{3}{c}{\textbf{Passenger Clips}} \\
\cmidrule(lr){2-3}\cmidrule(lr){4-6}\cmidrule(l){7-9}
\textbf{Round} & \textbf{Stops} & \textbf{Clips} & \textbf{Prec.} $\uparrow$ & \textbf{Rec.} $\uparrow$ & \textbf{F1} $\uparrow$ & \textbf{Prec.} $\uparrow$ & \textbf{Rec.} $\uparrow$ & \textbf{F1} $\uparrow$ \\
\midrule
\multicolumn{9}{@{}l}{\textbf{C3\_1} \quad \textit{GT: 67 stops, 159 passenger clips}} \\
\midrule

R1 & 92 & 283 & 0.663 & \textbf{0.910} & 0.767 & 0.505 & \textbf{0.899} & 0.647 \\
R2 & 66 & 166 & \textbf{0.894} & 0.881 & \textbf{0.887} & \textbf{0.687} & 0.717 & \textbf{0.702} \\
\midrule
\multicolumn{9}{@{}l}{\textbf{C3\_3} \quad \textit{GT: 24 stops, 91 passenger clips}} \\
\midrule

R1 & 41 & 132 & 0.561 & \textbf{0.958} & 0.708 & 0.636 & \textbf{0.923} & 0.753 \\
R2 & 25 & 98 & \textbf{0.880} & 0.917 & \textbf{0.898} & \textbf{0.816} & 0.879 & \textbf{0.847} \\
\bottomrule
\end{tabular}
\end{table*}

\subsection{Spatio-Temporal Grounded Payment Classification}

The retained clip provides passenger-level temporal grounding, but payment evidence may appear only briefly within a small farebox region. We therefore introduce a two-stage coarse-to-fine procedure for payment spatiotemporal grounding.

\paragraph{Shared farebox grounding.}
Let $P_{k_0,j_0}$ be the first retained passenger clip and $P_{k_0,j_0}[0]$ its first frame. SAM 3 localizes the farebox using the concept prompt $q_f=$ ``farebox payment box'':
\begin{equation}
B_f=\Psi_{\mathrm{SAM3}}(P_{k_0,j_0}[0],q_f)
\end{equation}
where $B_f=(x_1,y_1,x_2,y_2)$ contains normalized corner coordinates and is cached for subsequent clips. Stage-specific margins produce $B_f^{(1)}$ and $B_f^{(2)}$. Stage 1 uses $(0.35,0.35,-0.50)$ for broader context, whereas Stage 2 uses $(0,0.10,-0.50)$ for tighter localization. The negative bottom adjustment removes the lower portion of the detected box.

\paragraph{Stage 1 coarse classification and evidence selection.}
We uniformly sample 16 frames over the complete clip, crop each with $B_f^{(1)}$, and arrange them chronologically in a $4\times4$ contact sheet $G_{k,j}^{(1)}$. The first VLM call returns
\begin{equation}
    (\tilde y_{k,j},\tilde c_{k,j},\tilde\rho_{k,j},
    \mathcal{E}_{k,j})
    =\Phi_{\mathrm{pay}}(G_{k,j}^{(1)},q_{\mathrm{pay}})
\end{equation}
where $\Phi_{\mathrm{pay}}$ is the VLM and $q_{\mathrm{pay}}$ defines the five labels in $\mathcal{Y}$ together with their visual rules. The outputs $\tilde y_{k,j}$, $\tilde c_{k,j}$, and $\tilde\rho_{k,j}$ are the provisional label, confidence, and rationale. The key design here is to ask the VLM to select the evidence frame set $\mathcal{E}_{k,j}\subseteq{1,\ldots,16}$ that supports its reasoning, which reveals the inner cues of how the VLM makes its judgment.

\paragraph{Stage 2 grounded refinement and final classification.}
For clearer inspection of the key evidence, the earliest and latest evidence frames define a refined temporal interval for further reasoning. If Stage 1 returns no valid evidence frame, the full clip is used. We then uniformly resample 16 frames from the selected interval, crop them with $B_f^{(2)}$, and construct $G_{k,j}^{(2)}$. The second VLM call applies the same visual rules to the refined evidence
\begin{equation}
    (\hat y_{k,j},\hat c_{k,j},\hat\rho_{k,j})
    =\Phi_{\mathrm{pay}}(G_{k,j}^{(2)},q_{\mathrm{pay}})
\end{equation}
where $\hat y_{k,j}\in\mathcal{Y}$, $\hat c_{k,j}$, and $\hat\rho_{k,j}$ denote the final payment label, confidence, and rationale. Stage 1 searches the full passenger clip, whereas Stage 2 concentrates temporal sampling and spatial resolution on the selected evidence. The final decision is thus grounded in both the passenger clip and the shared farebox geometry, without requiring payment-specific model training.

\section{Experiments}

\subsection{Dataset and Implementation Details}
We collect 436 minutes of real onboard bus surveillance video, including two video records: C3\_1 was recorded during daytime operation from 8 a.m. to 2 p.m., while C3\_3 was recorded from 6 p.m. to 9 p.m. with nighttime scenes accounting for nearly half of the footage. Both videos have a resolution of $1280\times720$ at 10 FPS. We manually annotate every stop interval and passenger clip with its payment label. In our edge-cloud implementation, lightweight edge models monitor the door, track passengers, and segment clips, while cloud VLMs process only the grounded passenger clips and farebox contact sheets. We use GPT-4o for behavior classification and GPT-5.4-mini for payment classification. All experiments run on one NVIDIA RTX 4090 GPU with 24 GB of memory.

\begin{table*}[t]
\centering
\caption{Stage-wise payment classification results on C3\_1 and C3\_3. Prec. and Rec. denote class-wise precision and recall, respectively. Arrows in Stage 2 indicate changes relative to Stage 1.}
\label{tab-vlm-payment-results}
\footnotesize
\renewcommand{\arraystretch}{0.8}
\setlength{\tabcolsep}{7pt}
\begin{tabular}{@{}l l ll ll ll ll ll@{}}
\toprule
& \textbf{Overall} & \multicolumn{2}{c}{\textbf{Cash}} & \multicolumn{2}{c}{\textbf{QR}} & \multicolumn{2}{c}{\textbf{Swipe}} & \multicolumn{2}{c}{\textbf{Tap}} & \multicolumn{2}{c}{\textbf{Evade}} \\
\cmidrule(lr){3-4}\cmidrule(lr){5-6}\cmidrule(lr){7-8}\cmidrule(lr){9-10}\cmidrule(l){11-12}
\textbf{Video / Stage} & \textbf{Acc.} & \textbf{Prec.} & \textbf{Rec.} & \textbf{Prec.} & \textbf{Rec.} & \textbf{Prec.} & \textbf{Rec.} & \textbf{Prec.} & \textbf{Rec.} & \textbf{Prec.} & \textbf{Rec.} \\
\midrule
C3\_1 / Stage 1 & 0.349 & 0.400 & 0.455 & 0.185 & 0.556 & 0.062 & 0.091 & 0.400 & 0.196 & 0.867 & 0.491 \\
C3\_1 / Stage 2 & 0.313$\downarrow$ & 0.480$\uparrow$ & 0.545$\uparrow$ & 0.187$\uparrow$ & 0.778$\uparrow$ & 0.000$\downarrow$ & 0.000$\downarrow$ & 0.280$\downarrow$ & 0.137$\downarrow$ & 0.905$\uparrow$ & 0.358$\downarrow$ \\
\midrule
C3\_3 / Stage 1 & 0.485 & 0.400 & 0.308 & 0.324 & 0.750 & 0.333 & 0.056 & 0.407 & 0.688 & 0.950 & 0.559 \\
C3\_3 / Stage 2 & 0.536$\uparrow$ & 0.571$\uparrow$ & 0.308 & 0.381$\uparrow$ & 0.500$\downarrow$ & 0.400$\uparrow$ & 0.333$\uparrow$ & 0.429$\uparrow$ & 0.750$\uparrow$ & 0.846$\downarrow$ & 0.647$\uparrow$ \\
\midrule 
\midrule
\multicolumn{12}{c}{\textbf{Binary Evade/Non-Evade Classification}} \\
\midrule
\textbf{Video / Stage} & \textbf{Acc.} & \multicolumn{2}{c}{\textbf{Evade Prec.}} & \multicolumn{2}{c}{\textbf{Evade Rec.}} & \multicolumn{2}{c}{\textbf{Non-Evade Prec.}} & \multicolumn{2}{c}{\textbf{Non-Evade Rec.}} & \multicolumn{2}{c}{} \\
\midrule
C3\_1 / Stage 1 & 0.813 & \multicolumn{2}{c}{0.867} & \multicolumn{2}{c}{0.491} & \multicolumn{2}{c}{0.801} & \multicolumn{2}{c}{0.965} & \multicolumn{2}{c}{} \\
C3\_1 / Stage 2 & 0.783$\downarrow$ & \multicolumn{2}{c}{0.905$\uparrow$} & \multicolumn{2}{c}{0.358$\downarrow$} & \multicolumn{2}{c}{0.766$\downarrow$} & \multicolumn{2}{c}{0.982$\uparrow$} & \multicolumn{2}{c}{} \\
\midrule
C3\_3 / Stage 1 & 0.837 & \multicolumn{2}{c}{0.950} & \multicolumn{2}{c}{0.559} & \multicolumn{2}{c}{0.808} & \multicolumn{2}{c}{0.984} & \multicolumn{2}{c}{} \\
C3\_3 / Stage 2 & 0.837 & \multicolumn{2}{c}{0.846$\downarrow$} & \multicolumn{2}{c}{0.647$\uparrow$} & \multicolumn{2}{c}{0.833$\uparrow$} & \multicolumn{2}{c}{0.938$\downarrow$} & \multicolumn{2}{c}{} \\
\bottomrule
\end{tabular}%

\end{table*}

\begin{figure*}[t]
  \centering
  \includegraphics[width=0.9\textwidth]{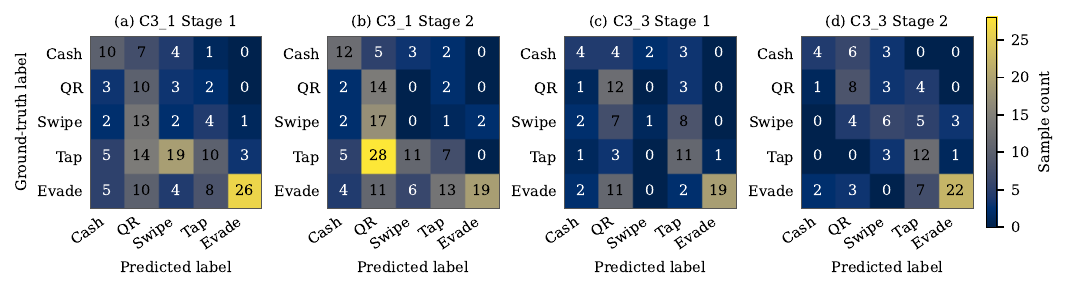}
  \caption{Payment-type confusion matrices for both VLM stages on C3\_1 and C3\_3. Each cell reports the number of samples from its ground-truth row assigned to the predicted column.}
  \label{fig-payment-confusion}
\end{figure*}

\subsection{Evaluation of Edge Modules}

We evaluate stop monitoring, passenger tracking, and passenger segmentation over the complete timelines. Predictions are matched one to one with ground truth at temporal IoU $\geq0.2$. At this stage, we obtain results from two rounds. Round 1 includes all detected stops and rule-based passenger clips. Round 2 removes non-boarding clips and merges neighboring stops when their retained clips are separated by less than 15 seconds. As shown in Figure~\ref{fig-edge-timelines} and Table~\ref{tab-edge-results}, Round 1 achieves high recall but overpredicts both stops and passenger clips because it retains non-boarding passengers and fragmented intervals. Round 2 brings the predicted counts close to the ground truth and removes many isolated intervals. Stop F1 increases from 0.767 to 0.887 on C3\_1 and from 0.708 to 0.898 on C3\_3, while passenger-clip F1 rises from 0.647 to 0.702 and from 0.753 to 0.847. 

This precision gain comes at the cost of lower stop-level and passenger-level recall. The primary bottleneck is the wrong removal of true boarding, resulting in an inherent precision-recall trade-off. Manual inspection further attributes the remaining errors to atypical activities, including the driver leaving the cockpit (at around 6,000 seconds in C3\_1), simultaneous boarding events, and passengers reappearing in the scene, etc. Direct sunlight in C3\_1 also causes image blur and a darkened bus interior, further degrading tracking accuracy.

\subsection{Evaluation of VLM Modules}

The VLM-based payment classification results are presented in Table~\ref{tab-vlm-payment-results} and Figure~\ref{fig-payment-confusion}. Overall, the system performs better on C3\_3 than on C3\_1, consistent with the edge-module results. C3\_1 exhibits more challenging illumination, with direct sunlight and shadows reducing the visibility of passengers and farebox interactions. In contrast, C3\_3 has more consistent lighting, enabling the VLM to recognize payment behaviors more reliably.

On C3\_3, the two-stage VLM improves five-class accuracy from 0.485 to 0.536, with higher recall for Swipe, Tap, and Evade. In contrast, Stage 2 reduces five-class accuracy on C3\_1 from 0.349 to 0.313. For binary evasion detection, C3\_1 accuracy decreases from 0.813 to 0.783, while Evade recall drops from 0.491 to 0.358. On C3\_3, accuracy remains unchanged at 0.837, but Evade recall increases from 0.559 to 0.647. These results indicate that, although Stage 2 provides more focused evidence, its effectiveness still depends strongly on visual quality.

The confusion matrices further reveal different refinement effects across the two videos. On C3\_1, Stage 2 concentrates more errors in the QR column, whereas on C3\_3 it reduces confusion involving QR and Tap. This difference explains why refinement benefits the visually more consistent C3\_3 sequence but not C3\_1. The VLM also struggles to distinguish QR from Tap because of their visual similarity, and Swipe is particularly difficult to recognize due to the small swipe region and subtle motion. Nevertheless, refinement increases Swipe recall on C3\_3 from nearly zero to 0.333, suggesting that more focused evidence can help the VLM capture fine-grained payment actions.

\section{Conclusions}

We propose GHR-VLM, a grounded hybrid framework that combines lightweight model-based perception with selective VLM reasoning for zero-shot transit video analytics. Model-based modules extract structured spatiotemporal passenger evidence, enabling VLMs to classify payment behavior without payment-specific training. The resulting pipeline transforms a single onboard surveillance stream into stop-, passenger-, and payment-level events. Experiments on real-world bus videos demonstrate the value of explicit grounding, although fine-grained payment recognition remains challenging. Future work will strengthen VLM reasoning and robustness under degraded video conditions.

\section*{Acknowledgment}

This work is supported by the NVIDIA Academic Program and the Transit IDEA Program. The authors gratefully acknowledge the Capital District Transportation Authority (CDTA) for providing the data used in this study and for its support of this research.

\bibliographystyle{IEEEtran}
\bibliography{main}

\end{document}